\documentclass{article} 
\usepackage{nips15submit_e,times}
\usepackage{url}
\usepackage{amsmath}
\usepackage{amssymb}

\newif\ifpdf
\ifx\pdfoutput\undefined
  \pdffalse                     
  \usepackage[dvips]{graphicx}
\else
  \pdfoutput=1                  
  \pdftrue
  \usepackage[pdftex]{graphicx}
\fi

\title{Tree-Cut for Probabilistic Image Segmentation}

\author{
Shell X.~Hu\thanks{SH conducted the work for this paper when he was at Oregon State University.} \\
Universit\'e Paris-Est, LIGM\\
\'Ecole des Ponts ParisTech\\
\texttt{hus@imagine.enpc.fr} \\
\And
Christopher K. I. Williams\\
School of Informatics\\
University of Edinburgh\\
\texttt{ckiw@inf.ed.ac.uk} \\
\And
Sinisa Todorovic\\
School of Electrical Engineering and Computer Science\\
Oregon State University\\
\texttt{sinisa@eecs.oregonstate.edu} \\
}

\newcommand{\be}{\begin{equation}}
\newcommand{\ee}{\end{equation}}
\newcommand{\bi}{\begin{itemize}}
\newcommand{\ei}{\end{itemize}}
\newcommand{\bea}{\begin{eqnarray}}
\newcommand{\eea}{\end{eqnarray}}

\newcommand{\bfmu}{\boldsymbol{\mu}}

\newcommand{\bfy}{\mathbf{y}}
\newcommand{\bfz}{\mathbf{z}}

\newcommand{\cut}[1]{}

\newcommand{\barAT}{A^{-}_T}
\newcommand{\YDi}{Y_{\Delta i}}
\newcommand{\gpb}{\emph{gPb}-owt-ucm}
\nipsfinalcopy 

\begin{document}

\maketitle

\begin{abstract}
This paper presents a new probabilistic generative model for image segmentation,
i.e.\ the task of partitioning an image into homogeneous regions.
Our model is grounded on a mid-level image representation, called a
region tree, in which regions are recursively split into subregions
until superpixels are reached. Given the region tree, image
segmentation is formalized as sampling cuts in the tree from the model.
Inference for the cuts is exact, and 
formulated using dynamic programming. 
Our tree-cut model can be tuned to sample segmentations at a
particular scale of interest out of many possible multiscale image
segmentations. This generalizes the common notion that there should be
only one correct segmentation per image. Also, it allows moving
beyond the standard single-scale evaluation, where the segmentation result for
an image is \emph{averaged} against the corresponding set of coarse
and fine human annotations, to conduct a scale-specific
evaluation. Our quantitative results are comparable to those of the
leading \gpb\ method, with the notable advantage that we additionally
produce a distribution over all possible tree-consistent segmentations of the image.
\end{abstract}

The image segmentation problem can be formalized as partitioning an
image into a set of non-overlapping regions so that each region
is (in some sense) homogeneous. Mathematically, this can be expressed
in terms of an objective function 
\begin{equation}
J(R) = \sum_{i \in R} \ell(Y_i) - c(R), \label{eq:J}
\end{equation}
where $R$ denotes the set of regions, $\ell(Y_i)$ is the 
log likelihood of the image data $Y_i$ that belongs to region $i$,
and $c(R)$ is a penalty term that e.g.\ penalizes the number 
of regions used. 

\begin{figure}
	\begin{minipage}[b]{0.25\linewidth}
	\centering
	\includegraphics[width=1.0\linewidth]{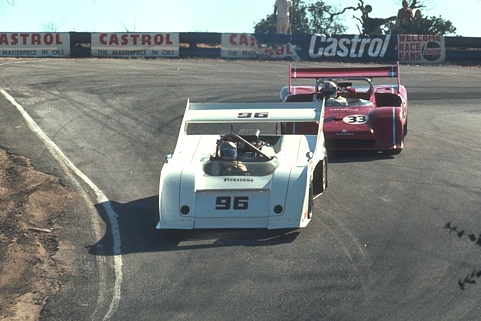} \\
    (a) \\
	\par\vspace{0pt}
	\end{minipage}\hfill%
	\begin{minipage}[b]{0.37\linewidth}
	\centering
	\includegraphics[width=1.0\linewidth]{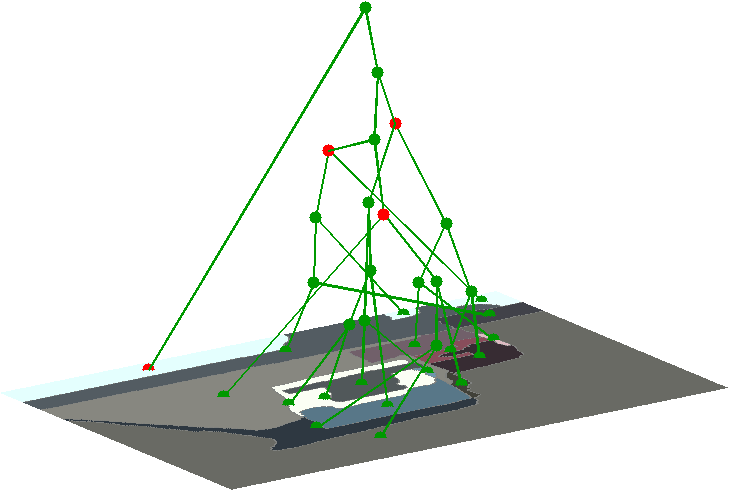} \\
	(c) \\
	\par\vspace{0pt}
	\end{minipage}\hfill%
	\begin{minipage}[b]{0.37\linewidth}
	\centering
	\includegraphics[width=1.0\linewidth]{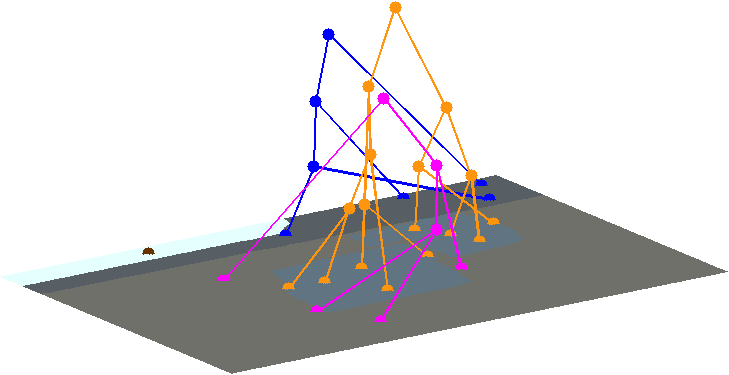} \\
	(e) \\
	\par\vspace{0pt}
	\end{minipage} \\
	\begin{minipage}[b]{0.25\linewidth}
	\centering
	\includegraphics[width=1.0\linewidth]{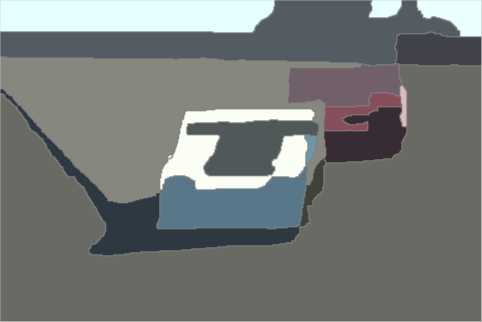} \\
    (b) \\
	\par\vspace{0pt}
	\end{minipage}\hfill%
	\begin{minipage}[b]{0.37\linewidth}
	\centering
	\includegraphics[width=1.0\linewidth]{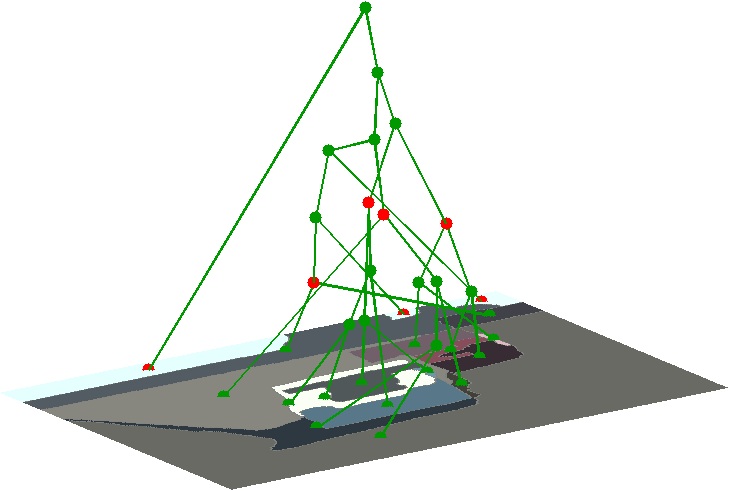} \\
	(d) \\
	\par\vspace{0pt}
	\end{minipage}\hfill%
	\begin{minipage}[b]{0.37\linewidth}
	\centering
	\includegraphics[width=1.0\linewidth]{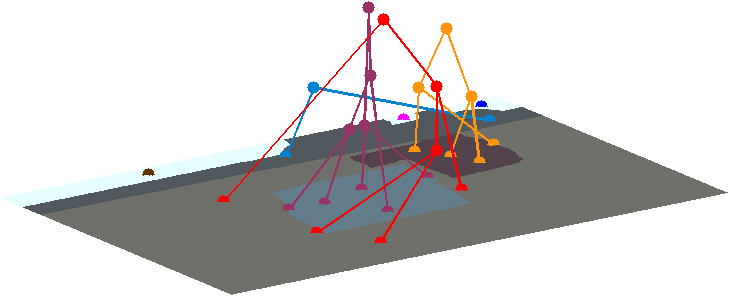} \\
	(f) \\
	\par\vspace{0pt}
	\end{minipage}
    \caption{Illustration of tree-cut (best viewed in color). (a) The original image. 
    (b) There are 18 superpixels corresponding to the leaves of the region tree.
    (c)-(d) The region trees showing the 
    active nodes (highlighted as red) after tree-cutting with different parameters. 
    (e)-(f) The resulting forests and segmentations. The nodes on the same subtree are rendered 
    with the same color. The coarse segmentation (e) has 4 regions,
and the fine segmentation (f) has 7 regions. These were obtained 
with $p=0.99$ and $p=0.90$ respectively, and $\lambda = 2 \times
10^{-4}$; the meaning  of these parameters is explained in section 
\ref{sec:model}.
    \label{fig:summary}}
\end{figure}

For input data with a one-dimensional ordering (e.g.\ a time series),
this problem is referred to as a \emph{changepoint problem}, and
dynamic programming can be used to find the optimal segmentation, see
e.g.\ \cite{killick-fearnhead-eckley-12}. However, such an approach is
not tractable in 2-d for image regions of arbitrary shape\footnote{One
  can use dynamic programming with rectangular regions defined by
  recursive splits, see e.g.\
  \cite{wang-pollak-wong-bouman-harper-siskind-06}.}.  Instead, given
a set of superpixels (i.e.\ an oversegmentation) of the image, we make
use of a region tree with superpixels as leaves, and with regions
being obtained as cuts in the tree. Given the region tree, we
formalize image segmentation as sampling from the posterior over cuts
in the tree from a generative model.  In this way, our image
segmentation amounts to selecting a segmentation from many
possible multiscale segmentations encoded by the region tree. Due to
the tree structure, MAP inference and sampling for the cuts is exact,
and formulated using dynamic programming.  An example of tree-cut in
operation is shown in Figure \ref{fig:summary}, with tuning for
coarse and fine segmentations. Some posterior
samples are shown in Figure \ref{fig:segs}.

The primary contribution of this paper is the formulation of a
principled and efficient probabilistic hierarchical model for image
segmentation.  Our model is generative, and thus allows for
probabilistic sampling of image segmentations at a particular scale of
interest, where the scale (e.g., coarse or fine) can be defined in
terms of a number of regions in the resulting
segmentation\footnote{Note that the number of regions in a
  segmentation, i.e.\ scale, is tightly related to their size. E.g.,
  large regions cannot occur in large numbers in the image.}. This can
be efficiently done by training only one parameter of our model on
ground-truth segmentations of a desired scale.  Our second
contribution is that we generalize the common notion that there should
be only one correct segmentation result per image. This allows us to
move beyond the evaluation methods of
\cite{arbaelaez-maire-fowlkes-malik-11} which \emph{average} their
result over the set of human annotations for a given image. In
contrast, we provide a more fine-grained evaluation, e.g.\ by tuning
our model to provide coarser or finer segmentations as per user
specifications during training.

The remainder of the paper is organized as follows: In section 
\ref{sec:model} we describe the model over tree-cuts, the likelihood
for regions, and  efficient computations for MAP inference and 
sampling from the posterior over segmentations. We also include a 
discussion of related work. In section \ref{sec:expts} we describe 
experiments on the BSDS datasets, including an evaluation of
tuning our method  to provide coarser or finer segmentations.
We conclude with a discussion and directions for future work.

\section{The Model \label{sec:model}}
We first describe the model over tree-cuts, and algorithms 
for MAP and sampling inference on the tree. We then discuss
the model used for the likelihood of image regions, how
parameters in the model can be set, and related work.

\subsection{Probabilities over tree cuts \label{sec:ptc}}
We start with a region tree $T$ whose leaves correspond to a
superpixel decomposition of the image. A region is defined by
\emph{activating} a node in the tree---this means that all of the
superpixels governed by that node are considered to form one region.
An active node corresponds to a \emph{cut} in the tree, in that it 
disconnects its subtree from the remainder of the tree.
A valid segmentation is obtained by activating a set of nodes
so that there is \emph{only one node active per path from leaf to
  root} (see e.g.\ \cite[sec~4.1]{silberman-sontag-fergus-14}),
which we refer to as the OONAPP constraint.

A convenient way of generating segmentations obeying the OONAPP 
constraint is in a top-down fashion. Starting at the root,
we can either activate that node (making the whole image one region),
or leave the root inactive and move down to consider its children,
by removing the edges between them and the root.
At each child we face the same question of whether to activate or
not. If this process continues down and we reach a leaf node then it must 
be activated, otherwise that superpixel would not be accounted 
for in the image.

We can ascribe probabilities to each configuration in the following
manner. Let each node $i$ have an activation probability $p_i$ and an
associated binary variable $z_i$, with prior probability $p(z_i = 1) =
p_i$.\footnote{As per the paragraph above, note that leaf nodes have their
$p$ variable set to 1.}
The $z$ variables are collected into a vector $\bfz$.  Let the
set of active nodes in $T$ be denoted as $A_T$, the nodes lying above
the active nodes as $\barAT$, and those below $A_T$ as $A^+_T$. The
$z$ variables are set to 1 for active nodes and those in $A^+_T$, and to
0 for nodes in $\barAT$. Then we have that
\begin{equation}
p(\bfz)
= \prod_{i \in \barAT} p(z_i = 0) \prod_{j \in A_T} p(z_j = 1)
\prod_{k \in A^+_T} \delta(z_k = 1). \label{eq:zA}
\end{equation}
Note that the nodes in $A^+_T$ are turned on with probability 1
given $A_T$. In relation to eq.\ (\ref{eq:J}), the penalty 
term $c(R)$  would be identified as $- \log p(\bfz|A_T)$. 

It is easy to show that this construction defines a valid probability
distribution over tree-cuts. Consider the subtree rooted at node $i$,
and assume that the distribution over configurations for the subtrees
are valid probability distributions. With probability $p_i$ we have
$z_i = 1$ (and all of the $z$ variables below it are also turned on,
as per eq.\ (\ref{eq:zA})). With probability $1-p_i$ we have $z_i = 0$
and the probability is evaluated as the product over the
configurations of the children. Let $\bfz^i$ denote $z_i$ and
all of the $z$ variables below it in the tree. To check that
$\sum_{\bfz^i} p(\bfz^i) =1$, break this sum up into sums over $z_i$
and over the variables in each subtree. By summing over $z_i$ 
and letting $\bfz^j$ denote the $z$ variables in the subtree rooted 
at node $j$ (a child of node $i$), we obtain 
\begin{equation}
\sum_{\bfz^i} p(\bfz^i) = p_i + (1-p_i) 
\prod_{j \in \mathrm{ch}(i)} \left( \sum_{\bfz^j} p(\bfz^j) \right)= 1
\label{eq:valid-distribution}
\end{equation}
based on the induction hypothesis that the subtrees are valid
probability distributions. The proposition in eq. (\ref{eq:valid-distribution}) holds by the base case, i.e.\
that the leaf nodes have their $p$ variables set to 1.

\subsection{Computing the total probability of the data summing over
  all configurations \label{sec:totalp}}
Let the likelihood of a region $i$ with node $i$ being
active be denoted as $p(Y_{\Delta i})$. At any non-leaf node
we can explain the data governed by it either by making
it the active node, or by passing the responsibility to 
its children. Thus the total probability $p(Y_i)$ of explaining
the data governed by node $i$ is given by the recursive formula
\begin{equation}
p(Y_i) = \sum_{\bfz^i} p(\bfz^i) p(Y_i|\bfz^i) = 
p_i p(\YDi) + (1-p_i) \prod_{j \in \mathrm{ch}(i)} p(Y_j) ,
\end{equation}
where as above we have decomposed the sum into terms over $z_i$ and
the remaining $z$ variables.
This recurison ``bottoms out'' for a leaf node $k$ with
$p(Y_k) = p(Y_{\Delta k})$, because $p_k=1$ for leaf nodes.
Thus if $Y$ denotes all of the image data, we can compute $p(Y)$ at
the root node by a bottom up pass through the tree. 

\subsection{Computing the MAP configuration, and
sampling from $p(\bfz|Y)$ \label{sec:map}}
We now wish to find a valid OONAPP configuration $\bfz^*$ that 
maximizes $p(\bfz^*|Y)$. Note that
\begin{equation}
p(\bfz|Y) = \frac{1}{p(Y)} \; p(\bfz) \prod_{j \in A_T(\bfz)}
p(Y_{\Delta_j}),
\end{equation}
where $p(\bfz)$ is defined as per eq.\ (\ref{eq:zA}).

Let $p^*(Y_i)$ denote the likelihood of the data governed
by node $i$ under the optimal choice of $z$ variables at and 
below node $i$\footnote{In the previous section we were summing
over paths, here we are maximizing, cf.\ a Viterbi alignment.}. 
At node $i$ we can either
explain all of the data in the subtree with probability $p_i p(\YDi)$
(the ``direct'' contribution), or
make use of the children (the ``indirect'' contribution). Thus 
\begin{equation}
p^*(Y_i) = \max [ p_i p(\YDi), \; (1-p_i) \prod_{j \in \mathrm{ch}(i)} p^*(Y_j)] .
\end{equation}
For leaf nodes $p^*(Y_k) = p(Y_{\Delta k})$.
Again we can compute $p^*(Y)$ by a bottom-up pass through the tree,
and obtain the optimal MAP configuration by a top-down backtracking pass having 
stored the argmax decisions made at each node in the tree.

\begin{figure}
	\centering
	\begin{minipage}[b]{0.245\linewidth}
	\centering
	\includegraphics[width=1.0\linewidth]{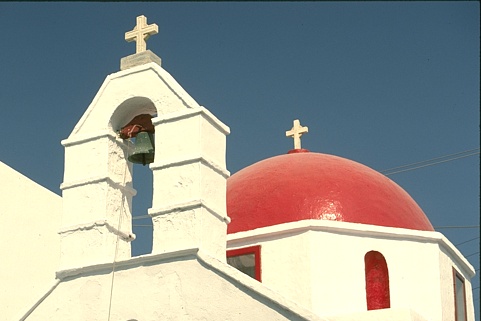}\\
	\includegraphics[width=1.0\linewidth]{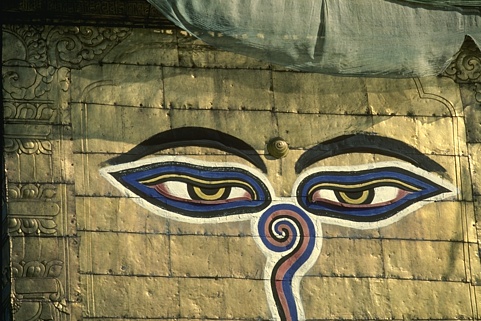}\\
	(a) \\
	\end{minipage}\hfill
	\begin{minipage}[b]{0.245\linewidth}
	\centering
	\includegraphics[width=1.0\linewidth]{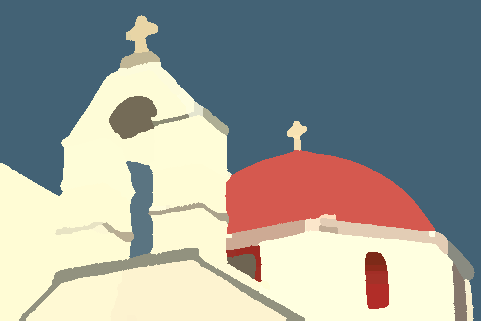}\\
	\includegraphics[width=1.0\linewidth]{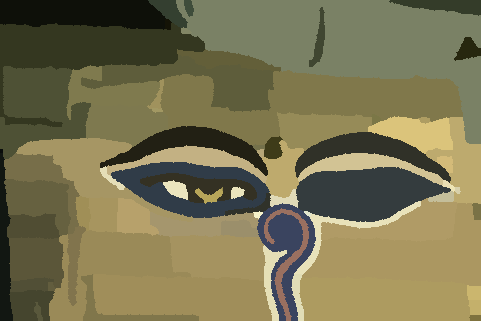}\\
	(b) \\                                 
	\end{minipage}\hfill
	\begin{minipage}[b]{0.245\linewidth}
	\centering
	\includegraphics[width=1.0\linewidth]{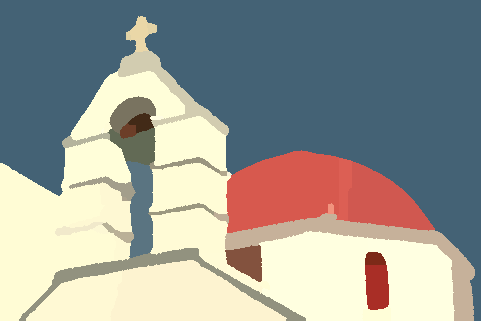}\\
	\includegraphics[width=1.0\linewidth]{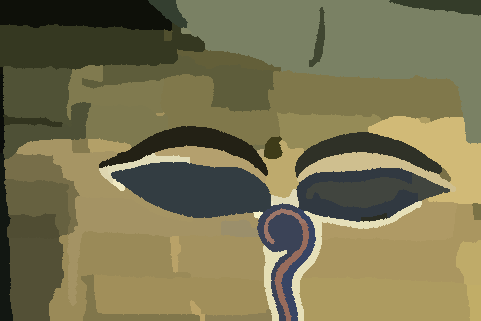}\\
	(c) \\                                 
	\end{minipage}\hfill
	\begin{minipage}[b]{0.245\linewidth}
	\centering
	\includegraphics[width=1.0\linewidth]{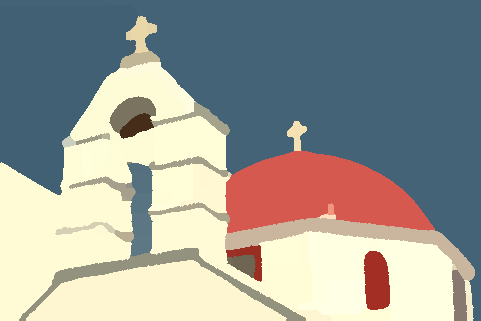}\\
	\includegraphics[width=1.0\linewidth]{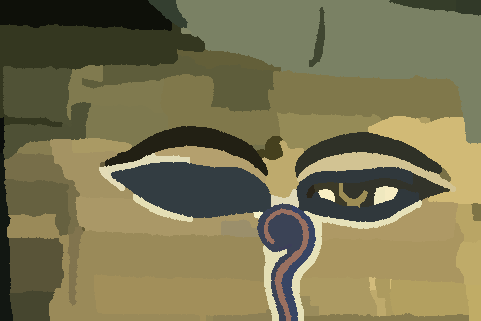}\\
	(d) \\                                               
	\end{minipage}\hfill
    \caption{Sampling from $p(\bfz|Y)$. (a) The original images which have  
    548 and 1263 superpixels respectively. (b)-(d) Samples with 39, 32 and 44 cuts for the first image and with 64, 64 and 69 cuts for the second image. The colors are
    selected as the median colors of the regions.
    \label{fig:segs}}
\end{figure}

{\bf Sampling from $p(\bfz|Y)$}. At the root we have 
computed the direct contribution $p_D(Y_r) = p_r p(Y_{\Delta r})$ and the
indirect contribution 
$p_I(Y_r) = (1-p_r) \prod_{j \in \mathrm{ch}(r)} p(Y_j)$. To sample over tree 
configurations we first sample the root with 
$p(z_r = 1|Y_r) = p_D(Y_r)/[p_D(Y_r) + p_I(Y_r)]$. If $z_r$ is active
we stop, alternatively we recurse into the subtrees, and 
sample there on the same basis. 
The variety of segmentations that can be obtained by sampling 
is illustrated in Figure \ref{fig:segs}. 

It is also possible to compute other quantities via message
passing on the tree, such as the marginals $p(z_i|Y)$.

\subsection{Likelihoods \label{sec:like}}
We need to define a likelihood model for data under a given node.
The simplest assumption is to assume that all pixels in the region
are drawn from a Gaussian distribution. Let there be 
pixels $\bfy^1, \ldots, \bfy^N$. These are $D=3$ dimensional vectors as they
are RGB values (other colour spaces are possible). The
maximum likelihood solutions for the mean and covariance matrix 
of this data are the empirical mean 
$\hat{\bfmu} = \frac{1}{N} \sum_{n=1}^N \bfy^n$ and 
$\hat{\Sigma} = \frac{1}{N}
\sum_{n=1}^N(\bfy^n-\hat{\bfmu})(\bfy^n-\hat{\bfmu})^T $. The
log likelihood of the data under this model is 
\begin{align*}
L & = - \frac{1}{2} \sum_{n=1}^N [ D \log (2 \pi)  + \log |\hat{\Sigma}| 
+ (\bfy^n - \hat{\bfmu})^T \hat{\Sigma}^{-1} (\bfy^n - \hat{\bfmu})]
\\
 & = - \frac{N}{2} [ D \log (2 \pi)  + \log |\hat{\Sigma}| ] - \frac{1}{2}
\mathrm{tr} [\hat{\Sigma}^{-1} \sum_n (\bfy^n - \hat{\bfmu})(\bfy^n -
\hat{\bfmu})^T ] \\
& = - \frac{N}{2} [ D \log (2 \pi)  + \log |\hat{\Sigma}| ] -
\frac{N}{2} \mathrm{tr}(I_D) 
 = - \frac{N}{2} [ D \log (2 \pi)  + \log |\hat{\Sigma}| + D] .
\end{align*}
Instead of maximizing over $\bfmu$ and $\Sigma$, it would be possible
to integrate them out via a Bayesian analysis; a Normal-Wishart prior
would be conjugate to the likelihood.
More complex likelihoods involving spatial correlations between pixels 
are discussed in sec.\ \ref{sec:discuss}.

In HMMs for speech recognition it is well-known
\cite{bahl-bakis-jelinek-mercer-80} that it can be useful to adjust 
the relative scaling of the likelihood and structural terms, as the
likelihood term arises from continuous observations, while the 
structural terms arise from discrete transition probabilities. 
In the recursive equations this is easily handled by simply
multiplying the log likelihood by a factor $\lambda > 0$.

In order to assess the effect of various likelihoods, we also
devised a ``ground-truth'' (GT) likelihood term, taking account 
of the segment labels in the annotations. Let $n_{ij}$ denote the
number of labels of type $j$ in region $i$, and let there be
$m$ different labels. We can ascribe a probability $\pi_{ij}$
for label $j$ in region $i$, and the log-likelihood under this
model is maximized by setting $\hat{\pi}_{ij} = n_{ij}/n_{i.}$
with $n_{i.} = \sum_k n_{ik}$. Thus we obtain a GT-likelihood of
\begin{equation}
L^{GT}_i = n_{i.} \sum_{j=1}^m \hat{\pi}_{ij} \log \hat{\pi}_{ij} .
\end{equation}
Note that this ``ground truth'' likelihood still omits spatial
correlations, so although it is very likely to be optimistic relative to 
the independent Gaussian model, it may not be an upper bound what
can be achieved.

\subsection{Setting the $p$ parameters \label{sec:setp}}
In the above formulation each internal node $i$ has an associated parameter
$p_i$; the corresponding $p$ for the leaves must be set to 1.
Note also that the region tree is different for each input
image. Thus it is not possible to learn the $p_i$ parameters directly
by accumulating information over images. 

One simple approach is to set a unique $p$ globally for a set of images and
segmentations, so as to optimize some measure of performance, such
as the covering score (see eq.\ (\ref{eq:covscore})). Note that 
maximization of the log likelihood $\log p(Y)$ wrt $p$ would
lead to setting $p=0$, as one can always obtain a higher likelihood
by having a specific model for each superpixel rather than merging
them. 

An interesting alternative would be to \emph{predict} $p_i$ 
based on information in the tree, such as the UCM scores 
(explained in sec.\ \ref{sec:relwork} below) of
a node, its children and its parent, and the fraction of nodes
in the tree that are governed by it. A suitable predictor would
be e.g.\ logistic regression. In this case the model would
be a hierarchical conditional random field (CRF), as the $p$
parameters would be set on the basis of the image data. We leave
this direction for future work.

\subsection{Related Work \label{sec:relwork}}
There are a vast number of papers on the topic of image
segmentation, going back decades. For a summary see e.g.\
\cite[ch.~5]{szeliski-10} and \cite[ch.~14]{forsyth-ponce-03}.
Generally these are based on: (a) Clustering of color and texture features
with mixture models, (b) Distribution-mode finding, (c) Graph partitioning, or 
(d) Boundary detection and closing the boundaries.

The work by Arbelaez et al.\ 
\cite{arbaelaez-maire-fowlkes-malik-11} is a notable example of boundary-detection based methods.
They first run the \emph{gPb} boundary detector to produce a boundary response map. The
detected boundaries are then closed using Oriented Watershed Transform (OWT), resulting in  
regions organized in an Ultrametric Contour Map (UCM). In the UCM each region is 
labelled with a weight value. By thresholding the UCM at a particular
weight, usually denoted by $k$, one obtains a segmentation at scale $k$. 
Varying $k$ produces a strict hierarchy of regions, which can be organized
in a region tree by their nesting properties. We refer to this work as the \emph{gPb}-owt-ucm method. 

It is worth noting that \emph{gPb}-owt-ucm {\em does not define a
  distribution over segmentations}, unlike our model.  A particular
value of $k$ (as obtained e.g.\ from a training set) selects one
segmentation in the tree. Other multiscale-segmentation approaches
also use a single threshold to identify the ``right'' scale of
segmentation (e.g.\
\cite{Segm_Compression_IJCV11,RegTree_Partitioning_CVPR08,
  ScaleInvarianceSegm_CVPR05}).

In general, one should make a distinction between image segmentation, which is an unsupervised
procedure, and semantic scene segmentation, where the aim is 
to label each (super)pixel with one of a discrete set of object class labels, e.g.\ 
sky, vegetation, vehicle etc. The ``Pylon model'' of 
\cite{lempitsky-vedaldi-zisserman-11} 
uses a notion of cuts in a region tree to define possible 
segment labellings, and then carries out inference in the tree to 
find the optimal cuts. Our cuts and the OONAPP constraint are
similar to their work, but our task -- image segmentation as opposed to semantic
segmentation -- is entirely different.

There has also been much prior work on using tree-structured 
models for tasks other than image segmentation. 
For example \cite{bouman-shapiro-94} used a quad-tree structured
belief network to carry out a semantic segmentation task; a
disadvantage of their approach is that the quad-tree structure does
not depend on the input image structure, and so can lead to ``blocky''
artifacts. The Dynamic Trees work of \cite{storkey-williams-03} aimed
to address this problem by using a prior distribution over many
candidate trees.  However, this meant that approximate variational
inference had to be used; in contrast we can use exact inference in
one data-driven tree. The Recursive Neural Network of
\cite{socher-lin-ng-manning-11} learns to merge agglomeratively a set
of superpixels to produce a tree structure. However, note that the
final segmentation depends only on the learned features applied on a
superpixel-by-superpixel basis.

\section{Experiments \label{sec:expts}}
Our experiments are carried out on the widely-used
Berkeley Segmentation Dataset (BSDS)
\cite{arbaelaez-maire-fowlkes-malik-11}.
There are two versions, the BSDS300 dataset (which consists of 200
training and 100 test images), and the BSDS500, which has 
an additional 200 test images, with the BSDS300 being used as
training/validation data. Each of the images was segmented by several
human annotators. 

To produce one segmentation for each image we carry out MAP inference,
as described in sec.\ \ref{sec:map}.

{\bf Evaluation metrics}: There are a number of ways to evaluate the
agreement between the output of a segmentation algorithm and the human
annotations.  Following \cite{arbaelaez-maire-fowlkes-malik-11} we
consider the segmentation covering score (COV), the variation of
information (VI) \cite{meila-05}, and the Probabilistic Rand Index
(PRI) \cite{unnikrishnan-pantofaru-hebert-07}. Notice that VI and the
Rand index were introduced for the evaluation of clusterings, i.e.\
they are not specialized to image segmentation (with its spatial
aspects).

The intersection-over-union (IoU) score between two regions $R$ and
$R'$ is defined as $\mathrm{IoU}(R,R') = |R \cap R'|/|R \cup R'|$. The covering
score of a segmentation $S$ by a segmentation $S'$ is then given by
\begin{equation}
COV(S \rightarrow S') = \frac{1}{N} \sum_{R \in S} |R| \max_{R' \in S'}
\mathrm{IoU}(R,R') .  \label{eq:covscore}
\end{equation}

When evaluating segmentations, Arbelaez et al.\ 
\cite{arbaelaez-maire-fowlkes-malik-11} consider two situations,
which they term optimal dataset scale (ODS) and optimal 
image scale (OIS). The difference is that their UCM scale 
parameter $k$ is set for the whole dataset in ODS, while
it is optimized on a per image basis (while averaging over the 
different segmentations) for OIS. 

We consider two experiments. In the first 
(sec.\ \ref{sec:bsdsstyle}) we use the same evaluation criteria as in 
\cite{arbaelaez-maire-fowlkes-malik-11}. Secondly, in sec.\
\ref{sec:candf} we show how the methods can be tuned on specific
sets of segmentations so as to produce methods that can e.g.\
provide coarser or finer segmentations.

\begin{table}
\begin{center}
BSDS300 \hspace*{3mm}
\begin{tabular}{|l|c|c||c|c||c|c|} \hline
& \multicolumn{2}{c|}{Covering} & \multicolumn{2}{c|}{PRI} &
\multicolumn{2}{c|}{VI} \\ \hline
& ODS & OIS & ODS & OIS & ODS & OIS \\ \hline
Human & 0.73 & 0.73 & 0.87 & 0.87 & 1.16 & 1.16 \\ \hline
\gpb  & {\bf 0.587} & 0.646 & {\bf 0.808} & 0.852 & 1.650 & 1.461 \\
Tree-cut    & 0.583 & {\bf 0.651} & 0.793 & {\bf 0.853} & {\bf 1.646} & {\bf 1.437} \\ \hline
Tree-cut (GT)  & 0.692 & 0.715 & 0.877 & 0.894 & 1.364 & 1.165 \\ \hline
\end{tabular}

\vspace*{3mm}

BSDS500 \hspace*{3mm}
\begin{tabular}{|l|c|c||c|c||c|c|} \hline
& \multicolumn{2}{c|}{Covering} & \multicolumn{2}{c|}{PRI} &
\multicolumn{2}{c|}{VI} \\ \hline
& ODS & OIS & ODS & OIS & ODS & OIS \\ \hline
Human & 0.72 & 0.72 & 0.88 & 0.88 & 1.17 & 1.17 \\ \hline
\gpb  & 0.587 & 0.647 & {\bf 0.826} & 0.856 & 1.687 & 1.469 \\
Tree-cut    & {\bf 0.592} & {\bf 0.650} & 0.810 & {\bf 0.857} & {\bf 1.633} & {\bf 1.448} \\ \hline
Tree-cut (GT)  & 0.688 & 0.707 & 0.881 & 0.890 & 1.391 & 1.280 \\ \hline
\end{tabular}
\end{center}
\caption{Results for the BSDS300 and BSDS500 datasets.
The Human and \gpb\ results are as in 
\cite{arbaelaez-maire-fowlkes-malik-11}, except that for 
\gpb\ we have re-calculated them to 3 dec pl. Between \gpb\ and TC
the best performing method is shown in {\bf bold}; for covering and PRI
higher numbers are better, while for VI lower is better. The 
tree-cut (GT) results use the ground-truth likelihood as described
in sec.\ \ref{sec:like}.
\label{tab:bsds}}
\end{table}

\subsection{BSDS-style evaluation \label{sec:bsdsstyle}}
The MAP results for the BSDS-style data evaluation are shown in Table
\ref{tab:bsds}, for both the BSDS300 and BSDS500 datasets. These
are results on the respective test sets, for ODS cases the parameters were
optimized on the training set. The results on covering, PRI and VI 
for \gpb\ and tree-cut (TC) are basically very similar,
with the numbers lying within about 1\% of each other. 
We also conducted experiments sampling 500 segmentations for 
a given image rather than computing the MAP; the mean scores
for here were very similar to the MAP result. To give some
idea of the variability, for the covering score
the average standard deviation (wrt the 500 samples) was around 0.01.

As would be expected, using the ``ground-truth'' likelihood (as
explained in sec.\ \ref{sec:like}) does improve the performance of
tree-cut markedly, which suggests that it could be beneficial to look
for better likelihood models (see also the discussion in sec.\
\ref{sec:discuss}).  We observe that the OIS measures for tree-cut GT
are very close to human performance, which makes sense given that it
is making use of the information in the human-provided segmentations.

For interest we report the parameters for the \gpb\ and TC models that 
maximize the ODS covering score.
For \gpb\ we searched over a grid of 100 equally-spaced 
$k$ values from $0.01$ to $1.00$. 
For our tree-cut method we considered scaling factors $\lambda$ for
the likelihood (see sec.\ \ref{sec:like})
from $10^{-4}$ to $10^{-3}$ in steps of $10^{-4}$. For $p$ we
considered 100 values between 0.0001 and 0.9999, with denser gridpoints 
towards the higher values (nearer to 1). The optimal values 
for BSDS300 were $k=0.170$, $p = 0.968$, $\lambda = 0.0004$. For
BSDS500 the optimal values were 
$k=0.170$, $p = 0.971$, $\lambda = 0.0005$.

\begin{table}
\begin{center}
\begin{minipage}{5.5cm}
\hspace*{2cm} \gpb \\\\
\begin{tabular}{|l|c|c|c|} \hline
Model & \multicolumn{3}{c|}{Data source} \\ \cline{2-4}
trained on & Coarse & Medium & Fine \\ \hline
Coarse & 0.620 & 0.562 & 0.539 \\ 
Medium & 0.589 & 0.572 & 0.564 \\
Fine   & 0.559 & 0.568 & 0.569 \\ 
All    & 0.589 & 0.572 & 0.564 \\ \hline
\end{tabular}  
\end{minipage} 
\hspace*{1cm}
\begin{minipage}{5.5cm}
\hspace*{2.3cm} Tree-cut \\\\
\begin{tabular}{|l|c|c|c|} \hline
Model & \multicolumn{3}{c|}{Data source} \\ \cline{2-4}
trained on  & Coarse & Medium & Fine \\ \hline
Coarse & 0.645 & 0.574 & 0.512 \\
Medium & 0.606 & 0.585 & 0.551 \\
Fine   & 0.540 & 0.561 & 0.561 \\ 
All    & 0.617 & 0.583 & 0.545 \\ \hline
\end{tabular}
\end{minipage}
\end{center}
\caption{Covering scores for the coarse/medium/fine experiments.
\label{tab:cvsf}}
\end{table}

\subsection{Coarser and finer segmentations \label{sec:candf}}
The scores in section \ref{sec:bsdsstyle} are obtained by
averaging over the set of segmentations available for a given
image. However, it is natural to want to be able to control the 
level of detail obtained, and this can be done in our method
by varying the $p$ parameter, and for \gpb\ by varying $k$.

To explore this aspect, we divided the BSDS500 images into three
groups, which we refer to as coarse, medium and fine
segmentations. Coarse images contain 1-8 segments, medium have 9-31,
and fine images contain 32 or more segments.  There are 168, 284 and
144 images in these classes respectively in the trainval set, and 99,
183 and 109 on the test set.

Models were trained (by grid search over the parameters as described
above) to optimize the covering score on one of the groups of training
data.  The test-set covering scores reported in in Table
\ref{tab:cvsf} were evaluated on the 200 BSDS500 test images.  For
both the TC and \gpb\ models it is notable that the performance is
better when training has taken place on the same type of data as is
used for testing (as seen by the diagonal elements in the tables 
being higher than other ones in the same column). The performance differences 
inside the columns are generally larger for the TC method than 
for \gpb; for example for testing on coarse data, the TC 
difference is 0.105, while the \gpb\ difference is 0.061.
This implies that the effect of tuning is stronger for TC than \gpb.
Comparing with the last line in the tables where either method 
was trained on all of the data (as in sec.\ \ref{sec:bsdsstyle}),
we see that it never hurts performance to train on a specific
subset of data (coarse/medium/fine) when testing on the same division.

For \gpb, the optimal $k$ values obtained were 0.25, 0.17 and 0.13
for coarse, medium and fine training respectively. These make
sense in that coarser segmentations correspond to higher UCM
values. For tree-cut for greater interpretability we have chosen to
fix $\lambda =0.0009$; the corresponding values of $p$ are $0.9999$,
$0.9963$ and $0.968$ resp. As would be expected coarser segmentations
lead to $p$ values nearer to 1 (as this gives a higher probability of
stopping at higher levels in the tree, see sec.\ \ref{sec:ptc})\footnote{The
scaling factor $\lambda$ interferes with the direct interpretability
of the $p$ parameters.}

\section{Discussion \label{sec:discuss}}
Our main contribution has been to introduce a principled and efficient
hierarchical model for image segmentation, extending the ideas from
1-d changepoint analysis to images. The results in sec.\
\ref{sec:bsdsstyle} show that this produces very similar results to
the \gpb\ method in terms of the standard BSDS evaluation scores using
the MAP segmentation. A great advantage is that our method does not
produce just one segmentation, but can sample over segmentations in
order to reflect different possible interpretations.

Our second contribution is to delve deeper into the image 
segmentation task, addressing the fact the BSDS data contains
multiple segmentations of a given image. We have shown in sec.\
\ref{sec:candf} how the tree-cut and \gpb\ methods can be 
tuned to produce coarser or finer segmentations, and 
have evaluated how this affects the segmentation covering 
scores. 

There are a number of directions in which this work can be
extended. Firstly, our model is generative, defining a distribution
$p(Y,\bfz)$ given a tree. It would be possible to create a 
discriminative version that aims to optimize $p(\bfz|Y)$
based on training set segmentations, in the spirit of 
\cite{silberman-sontag-fergus-14} (but note that their work
primarily addresses the semantic segmentation problem, not
image segmentation).

Secondly, the independent-pixel Gaussian likelihood
used in sec.\ \ref{sec:like} is very simple, and richer
models that take spatial correlations into account could be used.  For
example one could use a Gaussian Field of Experts model
\cite{weiss-freeman-07}, a Gaussian process model with special
features for modelling image textures
\cite{wilson-gilboa-nehorai-cunningham-14}, or a conditional
autoregressive (CAR) model, where each pixel is predicted on the basis
of some context (usually taken to be above and to the left).  However,
using such models would greatly increase the computational cost of
model fitting.

Thirdly, setting $p$ values globally in the tree is very simple, and
it would be very interesting to \emph{predict} these values from
information available in the tree, as discussed in sec.\
\ref{sec:setp}. This could also  mean selecting different values of in
different parts of the image, which would be useful for complex scenes
with large variations in object sizes (e.g., due to large differences
in their distances from the camera).
Finally, our method and \gpb\ both use a 
single, data-driven tree. However, it is natural to 
consider a distribution over possible trees, see e.g.\ the
work on Dynamic Trees in \cite{storkey-williams-03}. One way to 
incorporate this idea in the tree-cut model is to use 
a number of different trees $T_1, \ldots, T_m$, and sample segmentations from
each of them. Their relative weights could be obtained via the
likelihood $p(Y|T_i)$ as computed in sec.\ \ref{sec:totalp}.

\subsubsection*{Acknowledgments}
We thank Glencora Borradaile for helpful discussions. CW thanks 
Rich Zemel for a conversation about the importance of producing multiple
image segmentations. CW is an Associate Fellow of the 
Neural Computation and Adaptive Perception program of the
Canadian Institite for Advanced Research (CIFAR). 
This work was supported in part by grants NSF IIS 1302700 and NSF DEB 1208272.

\bibliographystyle{unsrt}
\bibliography{treecut15_arxiv}

\end{document}